\documentclass[letterpaper, 10 pt, conference]{ieeeconf}
\IEEEoverridecommandlockouts
\usepackage{amsmath}
\usepackage{amssymb}
\usepackage{amsfonts}
\usepackage{algorithm}
\usepackage{algpseudocode}
\usepackage{graphicx}
\usepackage{booktabs}
\usepackage{multirow}
\usepackage{caption}
\usepackage{makecell}
\usepackage[font=small]{caption}
\usepackage{dblfloatfix} 
\algnewcommand{\PARAMETERS}{\item[\textbf{Parameters:}]}

\usepackage[nolist, nohyperlinks]{acronym}
\acrodef{TFM}[TFM]{Tabular Foundation Models}
\acrodef{BO}[BO]{Bayesian Optimization}
\acrodef{RL}[RL]{Reinforcement Learning}
\acrodef{PBT}[PBT]{Population-Based Training}

\title{\LARGE \bf
Can Tabular Foundation Models Guide Exploration in \\ Robot Policy Learning?
}

\author{Buqing Ou$^{1}$ and Frederike Dümbgen$^{1,2}$
\thanks{$^{1}$Department of Mechanical Engineering, Carnegie Mellon University,
Pittsburgh, PA 15213, USA {\tt\small \{buqingo, fdumbgen\}@andrew.cmu.edu}. $^2$Part of this work was performed while Frederike Dümbgen was with Inria, Département d'informatique de l'ENS, CNRS, PSL Research University, Paris, France. This work was supported by the European Union, through the Horizon Europe research and innovation program under the Marie Skłodowska-Curie (GA no.101207106), and by the Department of Mechanical Engineering, Carnegie Mellon University.}
}%

\begin{document}

\maketitle
\thispagestyle{empty}
\pagestyle{empty}

\begin{abstract}
Policy optimization in high-dimensional continuous control for robotics remains a challenging problem. Predominant methods are inherently local and often require extensive tuning and carefully chosen initial guesses for good performance, whereas more global and less initialization-sensitive search methods typically incur high rollout costs. We propose \textbf{TFM-S3}, a tabular hybrid local-global method for improving global exploration in robot policy learning with limited rollout cost. We interleave high-frequency local updates with intermittent rounds of global search. In each search round, we construct a dynamically updated low-dimensional policy subspace via SVD and perform iterative surrogate-guided refinement within this space. A pretrained tabular foundation model predicts candidate returns from a small context set, enabling large-scale screening with limited rollout cost. Experiments on continuous control benchmarks show that TFM-S3 consistently accelerates early-stage convergence and improves final performance compared to TD3 and population-based baselines under an identical rollout budget. These results demonstrate that foundation models are a powerful new tool for creating sample-efficient policy learning methods for continuous control in robotics.
\end{abstract}

\section{INTRODUCTION}

\begin{figure}[t]
    \centering
    \includegraphics[width=\linewidth]{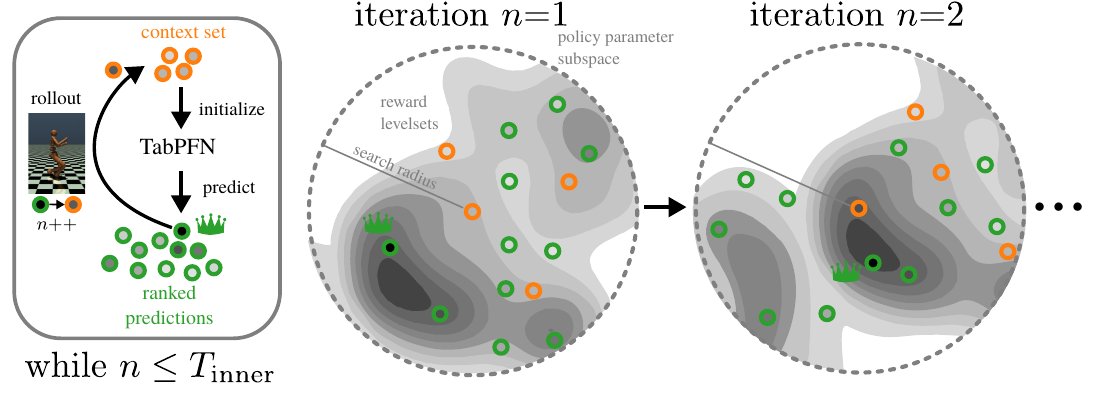}
    \caption{
\textbf{Tabular Foundation Model–guided Subspace Search.} The proposed method, TFM-S3, can be interleaved with any gradient-based policy training algorithm. It adds subspace-level global-search rounds as depicted in this figure. Within a dynamically updated policy parameter subspace, a set of candidate policies (colored circles) is generated. 
Only a small subset of candidates is evaluated through environment rollouts to form a context set (orange circles), which is used to initialize a Tabular Foundation Model.
The model then predicts the performance of the remaining candidates (green circles) and ranks them by predicted return.
The best candidate is evaluated and added to the context set before the next iteration.
This inner-loop procedure is repeated for $T_{\text{inner}}$ iterations, progressively guiding the search toward high-reward regions of the policy landscape with only limited rollout cost.
}
    \label{fig:abstract}
\end{figure}
Deep \ac{RL} has become a prominent approach for teaching robots new skills~\cite{sutton1998reinforcement}. However, as policy architectures scale to enable increasingly complex tasks, improving optimization performance remains a central challenge. A core difficulty lies in improving sample efficiency and training stability. 
Modern actor–critic methods, which are among the best performers in continuous control~\cite{schulman2017proximal,fujimoto2018addressing}, rely on neural policies with high-dimensional parameter spaces, often containing tens to hundreds of thousands of parameters~\cite{duan2016benchmarking}, resulting in highly non-convex optimization landscapes. While gradient-based methods scale to large parameter spaces, they perform inherently local updates and may converge slowly or stall in suboptimal regions~\cite{mnih2016asynchronous}. 
To improve exploration, population-based strategies such as evolution strategies and population-based training~\cite{salimans2017evolution} maintain diverse candidate policies, but require extensive rollout evaluations, resulting in substantial interaction cost.

\ac{BO} alleviates the need for extensive rollout evaluations by leveraging surrogate models to guide the selection of promising candidates~\cite{snoek2012practical}.
However, classical \ac{BO} becomes ineffective in high-dimensional parameter spaces due to data sparsity and the curse of dimensionality. When deep neural policies contain tens or even hundreds of thousands of parameters, constructing a reliable global surrogate from a limited number of samples becomes impractical.

However, a key observation is that effective policy updates often reside in low-dimensional structured manifolds~\cite{li2018measuring}. 
Restricting search to such subspaces reduces optimization complexity and makes surrogate modeling more feasible. 
At the same time, the relevant policy subspace evolves during policy training, inducing a non-stationary search geometry that violates the fixed-domain assumption underlying conventional \ac{BO}~\cite{shahriari2015taking}. This prevents the consistent aggregation of historical evaluations into a single global surrogate and calls for more flexible and adaptive surrogates.

Moreover, even within a single reduced subspace, accurate return prediction remains difficult when only a small number of rollout evaluations are available. Kernel-based models such as Gaussian processes~\cite{seeger2004gaussian} require careful hyperparameter tuning and often degrade in the presence of sparse data. 
In contrast, recent \ac{TFM}~\cite{hollmanntabpfn}, pretrained with strong inductive biases~\cite{hussain2025foundation}, provide robust small-sample generalization. 
When deployed within dynamically reconstructed policy subspaces, they enable reliable surrogate-guided candidate screening under limited rollout budgets, bridging gradient-based RL and sample-efficient surrogate-based search.

We propose TFM-S3 (Tabular Foundation Model–guided Subspace Search), a framework that interleaves gradient-based local updates with surrogate-guided global search in a dynamically reconstructed low-dimensional policy subspace.

In each search round, TFM-S3 first constructs a relevant low-dimensional policy subspace 
via an SVD-based procedure that captures the dominant update directions 
observed during recent training. 
Within each dynamically constructed policy subspace, TFM-S3 performs surrogate-guided global search under a limited rollout budget, as shown in Fig.~\ref{fig:abstract}: a small set of candidate policies and their real returns forms the context for a \ac{TFM}. The \ac{TFM} serves as the surrogate and predicts returns for additional candidates, allowing for a global screening of candidates without additional rollout cost. Only the best candidate is eventually chosen for rollout and added to the context set, and the process is repeated. This process is inspired by \ac{BO} with deterministic Thompson Sampling~\cite{daniel2018tutorial}, where the surrogate serves as the acquisition function to be maximized.
After this subspace-level global search, the next round of local updates is initialized by the best candidate found thus far, projected back to the original parameter space.

Our contributions are summarized as follows:
\begin{itemize}

\item We address the challenge of high-dimensional policy optimization by performing search in a dynamically updated low-dimensional subspace, constructed via a gradient-informed SVD procedure that adapts to the evolving training landscape.

\item We tackle the difficulty of surrogate modeling in RL by leveraging a pretrained \ac{TFM} which provides strong data-driven priors, enabling reliable return prediction without environment-specific tuning.

\item We reduce rollout requirements by combining subspace dimensionality reduction with surrogate-guided global search, requiring only a small number of real evaluations per search round while consistently accelerating convergence under identical rollout budgets.

\end{itemize}

Across continuous control benchmarks, TFM-S3 consistently accelerates early-stage convergence while also improving the final policy performance averaged across seeds, suggesting that structured surrogate-guided screening enhances solution quality with higher interaction efficiency than population-based methods.

\section{RELATED WORK}
\subsection{Zero-Order Methods in Reinforcement Learning}

Zero-order policy search methods optimize parameters directly through sampled rollouts without relying on gradient information. They are particularly appealing in settings where gradients are noisy or unstable.

\subsubsection{Population-Based and Evolutionary Methods}

Population-based approaches, such as evolutionary strategies and \ac{PBT}~\cite{jaderberg2017population}, improve exploration by evaluating multiple policy candidates in parallel. Compared to purely local updates, they explore a broader region of the parameter space. However, these methods typically require a large number of rollouts per iteration and rely on stochastic sampling~\cite{conti2018improving}. As policy dimensionality increases, maintaining adequate coverage of the search space becomes increasingly expensive due to the curse of dimensionality~\cite{hansen2001completely}, resulting in poor sample efficiency~\cite{omidvar2021review}.

\subsubsection{Bayesian Optimization}

\ac{BO} reduces rollout cost by learning a surrogate model of the return function and selecting candidates via acquisition functions~\cite{eriksson2019scalable}. This predictive guidance enables more sample-efficient search. Nevertheless, in high-dimensional neural policy spaces, surrogate models become inaccurate due to data sparsity~\cite{frazier2018bayesian}. As dimensionality grows, predictive quality deteriorates, limiting the effectiveness of conventional \ac{BO} for deep policy optimization in RL.

\subsection{Optimization in Low-Dimensional Subspaces}

To mitigate the curse of dimensionality in policy optimization, several works restrict search to a low-dimensional subspace, thereby improving sample efficiency and making surrogate modeling more tractable. Representative approaches include random low-dimensional embeddings for high-dimensional optimization~\cite{wang2016bayesian}, evolution strategies in structured subspaces~\cite{auger2012tutorial}, and gradient-informed subspace adaptation methods~\cite{chengcontinuous}. These methods typically rely on either fixed random projections or slowly adapting gradient-based directions.

In contrast, we dynamically reconstruct the policy subspace throughout training via an SVD-based procedure over recent policy updates, inspired by gradient-based dimensionality reduction techniques such as active subspace methods~\cite{constantine2015active}. As a result, the search directions explicitly track the dominant variations in the optimization trajectory, enabling a locally evolving coordinate system that adapts to gradient information rather than relying on a static or externally parameterized subspace. This dynamic reconstruction is particularly important in \ac{RL}, where the optimization landscape continuously changes as the policy and value functions co-adapt.

\subsection{Tabular Foundation Models}

Recent advances in tabular foundation models, particularly transformer-based pretrained regressors for tabular prediction~\cite{hollmann2025accurate}, have demonstrated strong performance in low-data regimes. 
Our work is inspired by GIT-BO~\cite{yugit}, which applies foundation-model-based surrogate search to Bayesian optimization in moderate-dimensional spaces (up to 500 dimensions). 
In contrast, we consider neural policy optimization in reinforcement learning, where the parameter space can easily exceed $10^5$ dimensions. 
This substantially higher dimensionality renders conventional surrogate modeling impractical in the full parameter space, motivating the use of dynamically constructed low-dimensional policy subspaces to enable efficient surrogate-guided search. 
Unlike GIT-BO, which derives search subspaces from gradients of the surrogate model with respect to the input parameters, we construct the subspace directly from recent policy-gradient directions via SVD, thereby allowing the search directions to adapt to the evolving optimization trajectory during reinforcement learning.

\section{Method}

\begin{figure}[t]
    \centering
    \includegraphics[width=\linewidth]{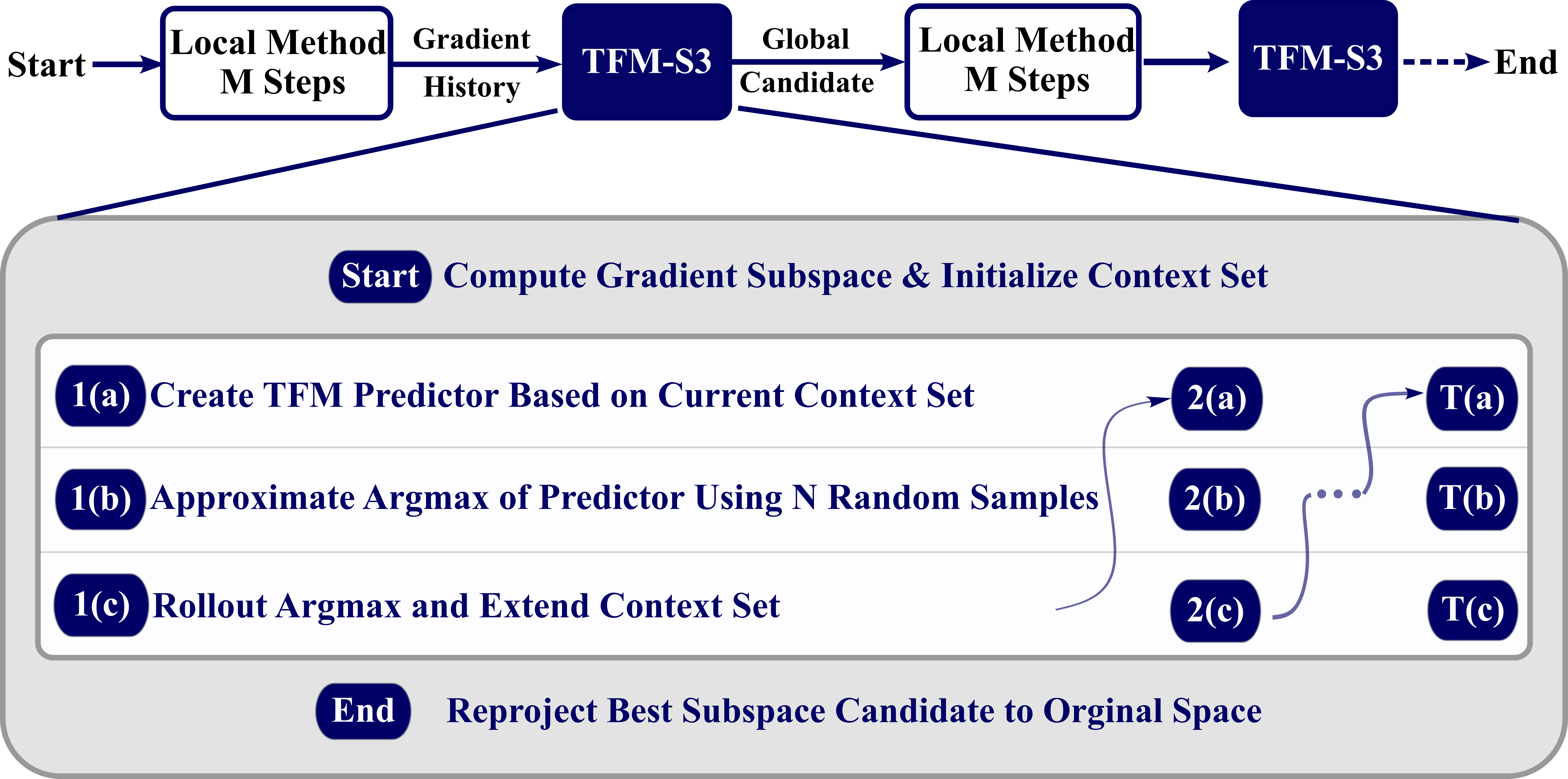}
    \caption{\textbf{Illustration of the proposed \textbf{TFM-S3} framework.} We interleave standard local reinforcement learning updates with subspace-level global search. For search rounds 1 to T, we use a tabular foundation model (TFM) based on a small context set to find the best candidate for rollout. The next local phase is initialized with the best candidate from this search phase, reprojected back to the original parameter space.}
    \label{fig:flow}
\end{figure}

\subsection{Problem Setting}

We consider \ac{RL} in continuous control tasks, where a stochastic policy $\pi_\theta$, parameterized with $\theta \in \mathbb{R}^D$, is trained to maximize the expected return $J(\theta)$ over a given time window.
Evaluating $J(\theta)$ requires collecting rewards by executing simulation rollouts or by interacting with a real environment. In either setting, limiting the training cost and time and, in particular, the sample complexity, \emph{i.e.}, the number of interactions with the system until convergence is achieved, is of foremost importance. 

As illustrated in Fig.~\ref{fig:flow}, our method interleaves local gradient-based RL updates with a periodic global search procedure (TFM-S3). 
The actor–critic backbone performs standard gradient descent updates, while every $M$ steps the TFM-S3 module is invoked to perform surrogate-guided exploration in a low-dimensional policy subspace.

Specifically, the procedure begins by constructing a gradient-based policy subspace and initializing a context set, as explained in Sec.~\ref{sec:subspace}. 
A tabular foundation model predictor is then constructed using the current context data (Sec.~\ref{sec:surrogate}), and the surrogate is used to find an optimal candidate in the subspace as explained in Sec.~\ref{sec:search}. 
The selected candidate is rolled out to obtain the true return and added to the context set. This process is repeated for $T$ surrogate-guided search steps, after which the best-performing subspace candidate is reprojected back to the original parameter space and used to update the actor.

\subsection{Dynamic Low-Dimensional Policy Subspace}\label{sec:subspace}

In global search round $t$, candidate policies are updated using
\begin{equation}\label{sub2whole}
\theta_{\mathrm{new}} = \theta_{\mathrm{base}}^{(t)} + A^{(t)} z,
\end{equation}
where $\theta_{\mathrm{base}}^{(t)} \in \mathbb{R}^D$ denotes the current base policy parameters, 
$A^{(t)} \in \mathbb{R}^{D \times r}$ is an orthonormal basis spanning the subspace capturing the dominant gradient directions, 
and $z \in \mathbb{R}^r$ represents the coordinates within this subspace.

We construct the subspace basis $A^{(t)}$ from recent local gradient information.
Specifically, we stack the most recent $Q$ gradient snapshots into a gradient matrix:
\begin{equation}\label{gradient_matrix}
G^{(t)} \;=\; \big[g_{t-Q+1}\;\; g_{t-Q+2}\;\; \cdots \;\; g_t\big]\in\mathbb{R}^{D\times Q}.
\end{equation}
We then compute a truncated SVD $G^{(t)} \approx U_r \Sigma_r V_r^\top$ and set
$A^{(t)} = U_r \in \mathbb{R}^{D\times r}$, the top-$r$ left singular vectors capturing
the dominant directions of variation in recent gradients.

This construction provides a concrete interpretation of the subspace coordinates.
For any $z \in \mathbb{R}^r$, the update $A^{(t)} z$ corresponds to a weighted linear combination of the dominant gradient directions captured by the subspace.
Optimizing over $z$ therefore amounts to searching for an optimal mixture of the most informative recent update directions.
This stands in contrast to arbitrary perturbations in the full parameter space $\mathbb{R}^D$, as in naive sampling-based methods, or to following only the local steepest descent direction (possibly with momentum), as in accelerated gradient descent.
As training progresses, the gradient distribution evolves; consequently, both $\theta_{\mathrm{base}}^{(t)}$ and $A^{(t)}$ are updated over time, resulting in a dynamically adapting search manifold.

\subsection{Surrogate Modeling}\label{sec:surrogate}

Surrogate-based optimization methods, such as Bayesian optimization, typically assume a fixed search domain and accumulate historical observations to update a global posterior model. However, in our setting, the search space is defined by the evolving $A^{(t)}$ and $\theta_{\mathrm{base}}^{(t)}$. As a result, maintaining a single global surrogate model becomes incompatible.

Instead, we adopt a \textit{localized} surrogate modeling strategy: within each subspace instance, we fit a predictor using only the evaluations collected during that global round. 
Concretely, we employ a \acf{TFM}. TFMs are pretrained on diverse regression tasks and provide strong small-sample generalization without task-specific hyperparameter tuning. At inference time, a \ac{TFM} is conditioned on the available input-output pairs, yielding a predictor that can be queried on new inputs to produce estimated outputs.

\subsection{Foundation Model--Guided Global Search}\label{sec:search}

Given the dynamic subspace defined at iteration $n$, we perform a structured search in the low-dimensional coordinate space $\mathbb{R}^r$. 

\subsubsection{Predictive Modeling in Subspace}
We propose a search method that limits the number of real rollouts by replacing them with predictions from a tabular foundation model and only rolling out promising candidates. This is in contrast with a population-style search procedure with interaction cost proportional to the number of screened candidates. 

We sample a small set of candidates by sampling isotropic Gaussian perturbations of variance $\sigma^2$ around a center $z_s$ and then projecting them to an $\ell_2$ trust region of radius $r_{\mathrm{local}}$:
\begin{equation}\label{sample}
\tilde z_i \sim \mathcal{N}(z_{\mathrm{s}},\,\sigma^2 I),\qquad
z_i \leftarrow \Pi_{\mathcal{B}_2(z_{\mathrm{s}}, r_{\mathrm{local}})}(\tilde z_i),
\end{equation}
where $\Pi_{\mathcal{B}_2}$ denotes projection onto the ball $\mathcal{B}_2(z_{\mathrm{s}}, r_{\mathrm{local}})=\{z:\|z-z_{\mathrm{s}}\|_2\le r_{\mathrm{local}}\}$. We call the function to sample $P$ points according to this procedure $\textsc{GaussianSample}(z_s, P)$.

For the initial context set, we choose $z_c=0$ and choose $K$ samples, leading to the sample set
\begin{equation}
    \mathcal{Z}_c^{(t)} = \textsc{GaussianSample}(0, K).
\end{equation}
We use these subspace samples to create the \emph{context set}:
\begin{equation}\label{form_context}
\mathcal{C}^{(t)} = \left\{ \left(z_j, J(\theta_\mathrm{base} + A^{(t)}z_j)\right) \right\}_{z_j\in\mathcal{Z}_c^{(t)}},
\end{equation}
where the returns $J(\cdot)$ are obtained through actual policy rollouts.

We construct a predictive model $\hat{J}^{(t)}:\mathbb{R}^r \to \mathbb{R}$ by initializing a \ac{TFM} with the context set $\mathcal{C}^{(t)}$, as explained in Sec.~\ref{sec:surrogate}.  This predictor estimates the return of a candidate policy based on its subspace coordinate $z$ without performing additional rollouts.

Next, we sample a set of $N \gg K$ candidate subspace coordinates around the best policy from the context set, $z^\star$:
\begin{equation}
    \mathcal{Z}_e^{(t)} = \textsc{GaussianSample}(z^\star, N),
\end{equation}
The predicted returns for the $N$ candidates are then computed and collected into the evaluation set
\begin{equation}\label{tabpfn_pred}
\mathcal{E} = \left\{\left(z_i, \hat{J}^{(t)}(z_i)\right)\right\}_{z_i\in\mathcal{Z}_e^{(t)}}.
\end{equation}
Finally, we obtain the most promising subspace candidate by selecting the candidate with the highest predicted return,
\begin{equation}\label{argmax}
(z^{*}, \hat{y}^*) = \arg\max_{(z_i,{y_i})\in\mathcal{E}} y_i.
\end{equation}

\subsubsection{Iterative Refinement}

Rather than performing a one-shot selection of the best candidate based on predicted rewards, we conduct a sequential refinement process within each search round.
Starting from an initial context set of size $K$, the surrogate-guided global search is repeated to produce high-quality samples.

At inner iteration $n$, the foundation model is conditioned on the current context set $\mathcal{C}_n^{(t)}$ to obtain an updated predictor $\hat{J}_n^{(t)}$. 
A new candidate pool of size $N$ is sampled according to \eqref{sample}, and predicted returns are computed using \eqref{tabpfn_pred}. The candidate with the highest predicted return is selected and evaluated via a real rollout. The context set is then augmented with the latest rollout, and the loop is repeated.

After $T$ refinement steps, the best evaluated candidate within the entire context set is selected to update the policy using \eqref{sub2whole}.

\begin{algorithm}[t]
\caption{Interleaving Gradient-Based RL with Subspace-Level Global Search \textsc{TFM-S3}.}
\label{alg:main}
\begin{algorithmic}[1]

\PARAMETERS Start step $M_{\text{start}}$, search period $M$, gradient window $Q$.

\State Initialize an off-policy actor--critic agent replay buffer $\mathcal{B}$.

\For{$m = 1,2,\ldots$ \textbf{(environment steps)}}
    \State Interact with the environment to collect transitions and store them in $\mathcal{B}$.
    \State Perform standard gradient-based RL updates (critic and actor updates) using mini-batches sampled from $\mathcal{B}$.
    \If{$m \ge M_{\text{start}}$ \textbf{and} $(m - M_{\text{start}}) \bmod M = 0$}
        \State Form $G^{(t)}$ from actor gradient history as in~\eqref{gradient_matrix}
        \State $\theta \leftarrow \textsc{TFM-S3}(\theta, G^{(t)})$
    \EndIf
\EndFor

\end{algorithmic}
\end{algorithm}

\vspace{2mm}
\begin{algorithm}[t]
\caption{\textsc{TFM-S3}: Tabular Foundation Model-Guided Subspace-Level Global Search.}
\label{alg:globalsearch}
\begin{algorithmic}[1]

\Require Current actor parameters $\theta_\mathrm{base}\in\mathbb{R}^D$, gradient history $G \in \mathbb{R}^{D \times Q}$

\PARAMETERS 
$r$ (subspace rank),
$T$ (inner iterations), 
$K$ (initial context size), 
$N$ (candidates per iteration), 
$r_{\text{local}}$ (sampling radius), 
$\sigma$ (sampling noise)

\State Compute truncated SVD $G^{(t)} \approx U_r \Sigma_r V_r^\top$ 
\State Set $A^{(t)} \leftarrow U_r \in \mathbb{R}^{D\times r}$
        
\State Choose points: $\mathcal{Z}_c \leftarrow \textsc{GaussianSample}(0, K)$ 
\State Initialize context: $\mathcal{C} \leftarrow \{(z_i, J(\theta_{\mathrm{base}} + Az_i))\}_{z_i\in\mathcal{Z}_c}$ 

\For{$n = 1,2,\ldots,T$}
    \State Get predictor: $\hat{J}(\cdot) \leftarrow$ foundation model on $\mathcal{C}$
    \State Find center: $(z_c, y_c) \leftarrow \arg\max_{(z,y)\in \mathcal{C}} y$
    \State Sample points: $\mathcal{Z}_e \leftarrow \textsc{GaussianSample}(z_c, N)$ 
    \State Evaluate points: $\mathcal{E} \leftarrow \{(z_i, \hat{J}(z_i))\}_{z_i \in \mathcal{Z}_e}$
    \State Choose best: $(z^\star, \hat{y}^\star) \leftarrow \arg\max_{(z, y)\in\mathcal{E}} y$
    \State Rollout best: $y^\star \leftarrow J(\theta_{\mathrm{base}} + A z^\star)$
    \State Extend context: $\mathcal{C} \leftarrow \mathcal{C} \cup \{(z^\star, y^\star)\}$
\EndFor

\State $(z_{\text{best}}, y_{\text{best}}) \leftarrow \arg\max_{(z_i,y_i)\in \mathcal{C}} y_i$
\State $\theta^\star \leftarrow \theta_{\mathrm{base}} + A z_{\text{best}}$

\State \textbf{return} $\theta^\star$

\end{algorithmic}
\end{algorithm}

\subsection{Conclusion}
Overall, TFM-S3 integrates dynamic subspace learning with foundation-model–guided global search in a unified framework. 
Local gradient descent ensures consistent policy improvement, while periodic subspace-based global search enables structured exploration beyond immediate gradient directions. 
By restricting candidate generation to a dynamically evolving low-dimensional manifold and leveraging a pretrained \ac{TFM} for iterative screening, the proposed approach achieves a global yet interaction-efficient policy.

\section{Results}
\subsection{Experimental Setup}
\begin{figure*}[t]
    \centering
    \includegraphics[width=\textwidth]{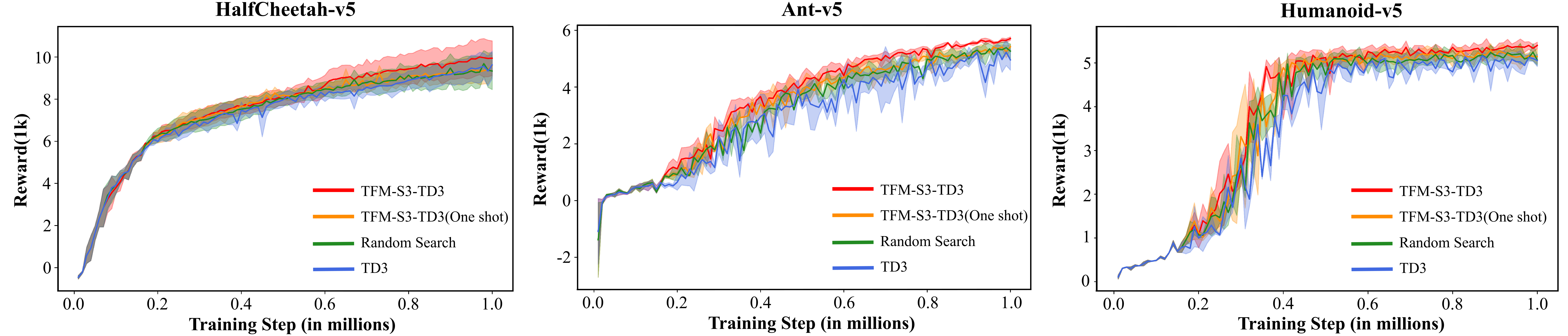}
    \caption{\textbf{Convergence plots showing added value of TFM-S3 plugged into TD3, compared with baselines.} We plot the learning curves of TFM-S3-TD3 (proposed), TFM-S3-TD3 (One shot), Random Search (32 candidates), and vanilla TD3 on HalfCheetah-v5, Ant-v5, and Humanoid-v5. We plot the cumulative reward as a function of training steps, averaged over 5 seeds. The shaded region corresponds to the standard deviation.}
    \label{fig:training_curves}
\end{figure*}

\begin{table*}[t]
\centering

\caption{\textbf{Quantitative performance comparison.}
Steps (\%) reports the fraction of the total training budget (1M environment steps) required for an algorithm to first reach a predefined target performance threshold. Specifically, for each training curve, we define a target return (0.9 of the final performance) and compute the earliest training step.}
\setlength{\tabcolsep}{2pt}
\renewcommand{\arraystretch}{1.0}

\small 

\begin{tabular}{lcccccccccccc}
\toprule
& \multicolumn{3}{c}{TFM-S3-TD3}
& \multicolumn{3}{c}{TFM-S3-TD3(One shot)}
& \multicolumn{3}{c}{Random Search TD3}
& \multicolumn{3}{c}{TD3} \\

\cmidrule(lr){2-4}
\cmidrule(lr){5-7}
\cmidrule(lr){8-10}
\cmidrule(lr){11-13}

Environment
& Mean & Std (\%) & Steps (\%)$\downarrow$
& Mean & Std (\%) & Steps (\%)$\downarrow$
& Mean & Std (\%) & Steps (\%)$\downarrow$
& Mean & Std (\%) & Steps (\%)$\downarrow$ \\
\midrule

HalfCheetah-v5
& \textbf{9863.4} & 40.5 & \textbf{62.7}
& 9576.9 & \textbf{37.7} & 68
& 9686.3 & 38.4 & 66
& 9624.1 & 39.6 & 71.3 \\

Ant-v5
& \textbf{5608.2} & \textbf{6.1} & \textbf{63}
& 5419.8 & 8.4 & 70.1
& 5386.9 & 20.8 & 69.2
& 5279.6 & 22.5 & 71.6 \\

Humanoid-v5
& \textbf{5350.9} & \textbf{7.4} & \textbf{38}
& 5271.2 & 16.6 & 43
& 5245.2 & 12.4 & 48
& 5127.6 & 13.7 & 48 \\

\bottomrule
\end{tabular}
\label{tab:results}
\end{table*}

\subsubsection{Environments and Training Protocol}

We evaluate the proposed method on three MuJoCo continuous-control benchmarks:
HalfCheetah-v5, Ant-v5, and Humanoid-v5.
All experiments are conducted with a total training budget of 1M environment steps
and are repeated over 5 random seeds.

We instantiate the proposed surrogate-guided subspace search framework on top of TD3, which serves as the underlying \ac{RL} backbone. The core TD3 update rules remain unchanged; the additional search layer only determines how candidate policy parameters are selected within each update cycle. Therefore, performance differences reflect improvements in the search mechanism.
Although TD3 is used in our experiments, the proposed surrogate-guided subspace search is orthogonal to the choice of backbone and can be integrated with other off-policy or on-policy algorithms, such as SAC or PPO. In our implementation, the \ac{TFM} is constructed using \textbf{TabPFN-v2}.

\subsubsection{Search Hyperparameters}

The surrogate-guided subspace search is activated after an initial warm-up phase of
$M_{\text{start}} = 150{,}000$ environment steps.
This delayed activation is motivated by the observation that, during very early training,
the policy landscape is highly non-stationary and the \ac{TFM} exhibits
low predictive reliability, as analyzed in Sec.~\ref{sec:prediction_accuracy}. Prediction quality improves substantially after the initial training phase.
Delaying the search allows the backbone policy to reach a more stable regime before surrogate-guided refinement is applied.

After the warm-up phase, a search is performed periodically every
$M = 10{,}000$ environment steps. Within each search round, the sampling radius $r_{\text{local}}$ is $0.01$ and the search noise
$\sigma$ is $0.005$.
We perform $T= 16$ sequential inner iterations.
The initial context set consists of $K = 16$ rollout-evaluated candidates (including the center point).
At each inner iteration, $N = 256$ candidate points are sampled in the subspace and evaluated with the foundation-model surrogate.
Only the best candidate is evaluated via a real rollout and added to the context set.
As a result, each search round uses a total rollout budget of 32 evaluations.

All search-related hyperparameters are fixed across environments.
No environment-specific hyperparameter tuning is performed.

\subsubsection{Subspace Configuration}

In our experiments, we fix the subspace dimension to $r = 15$ across all environments.
The three MuJoCo tasks considered in this work have comparable policy parameter dimensionality and similar gradient statistics, and we empirically observed that a moderate subspace rank provides a good balance between search expressiveness and rollout efficiency.
Fixing a shared rank across tasks simplifies the design and avoids environment-specific hyperparameter tuning.

\subsection{Performance Comparison on MuJoCo Benchmarks}

We compare the following four variants, the results are shown in Fig.~\ref{fig:training_curves} and Table~\ref{tab:results}:

\begin{itemize}
    \item \textbf{TFM-S3-TD3.} 
    Our full method. In each search round, we initialize a context set with 16 rollout-evaluated candidates, followed by 16 sequential surrogate-guided evaluations (one candidate per iteration). The final policy update is selected from the entire evaluated set.

    \item \textbf{Random Search TD3.} 
    A non-iterative baseline operating in the same dynamically learned subspace. At each round, 32 candidate perturbations are sampled and evaluated via rollouts, and the best-performing candidate is selected. This baseline uses the same rollout budget (32 evaluations per round) as our method but without surrogate-guided refinement.

    \item \textbf{TFM-S3-TD3 (One Shot).} 
    An ablation variant that removes the iterative refinement mechanism. Starting from 16 initially evaluated candidates, only a single surrogate-guided evaluation is performed before updating the actor. This isolates the contribution of multi-step internal refinement.

    \item \textbf{TD3.} 
    The standard TD3 algorithm without any additional global search layers.
\end{itemize}

All search-based variants operate in the same dynamically updated low-dimensional subspace, ensuring that comparisons isolate the effect of candidate selection strategy under identical geometric constraints.

Across all environments, \textbf{TFM-S3-TD3} consistently accelerates early-stage learning. The performance gap is particularly evident during the middle phase of training, where iterative surrogate-guided refinement enables more reliable identification of high-quality policy updates compared to purely random sampling.

Compared to \textbf{Random Search TD3}, which uses the same evaluation budget, our method achieves higher returns earlier in the training. Since both methods evaluate the same number of candidates per round within the same subspace, this improvement reflects more effective candidate screening rather than increased computational resources.

The \textbf{One Shot} variant further clarifies the importance of internal iteration. While it performs better than pure random search, it consistently underperforms the full iterative version. This indicates that sequential refinement—where surrogate predictions are repeatedly updated with newly evaluated points—plays a critical role in progressively improving candidate quality. We further investigate this in Sec.~\ref{sec:prediction_accuracy}.

Finally, compared to vanilla \textbf{TD3}, all subspace-based search variants exhibit faster transitions into high-performance regimes. However, the iterative surrogate-guided approach achieves the strongest final performance across all tasks, suggesting that structured internal search not only accelerates convergence but also improves policy quality.

In addition, the iterative variant exhibits reduced variability across random seeds, as evidenced by narrower performance bands. This suggests that surrogate-guided refinement stabilizes policy updates by reducing the probability of selecting suboptimal perturbations.

\subsection{Internal Search Dynamics Across Training}

\begin{figure}[t]
    \centering
    \includegraphics[width=1.0\linewidth]{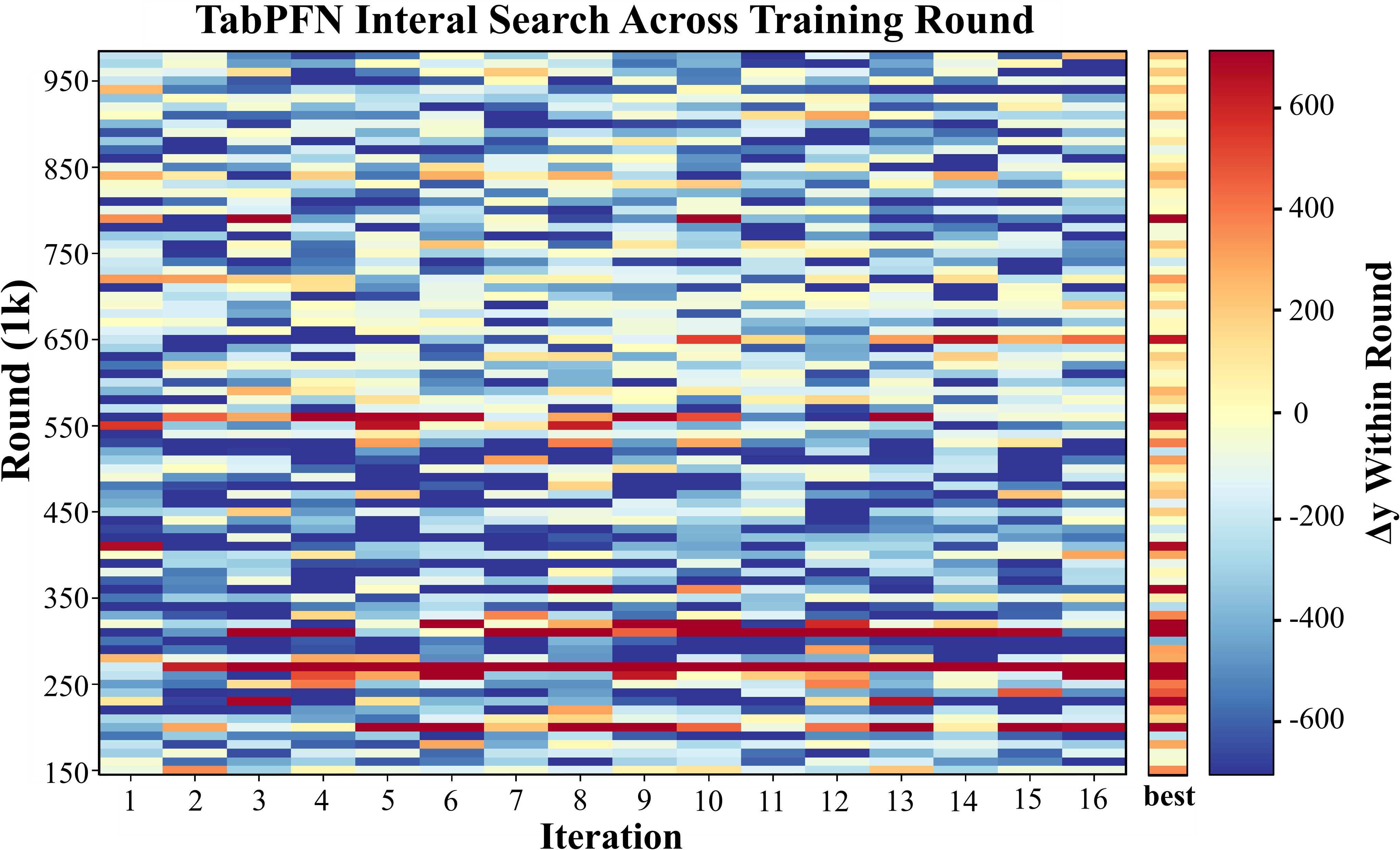}
    \caption{
\textbf{Improvement of global candidates across rounds.}
Each row corresponds to one TabPFN subspace-level global search round, labeled by training step, and each column corresponds to an inner iteration (1--16).
The color encodes the improvement value $\Delta y$, where positive values indicate candidate policies outperforming the baseline.
The narrow strip between the heatmap and the color bar, labeled \textbf{best}, shows the maximum improvement $\max_k \Delta y_{r,k}$ achieved within each round.
}
    \label{fig:heat_map}
\end{figure}

\begin{figure}[t]
    \centering
    \includegraphics[width=1.0\linewidth]{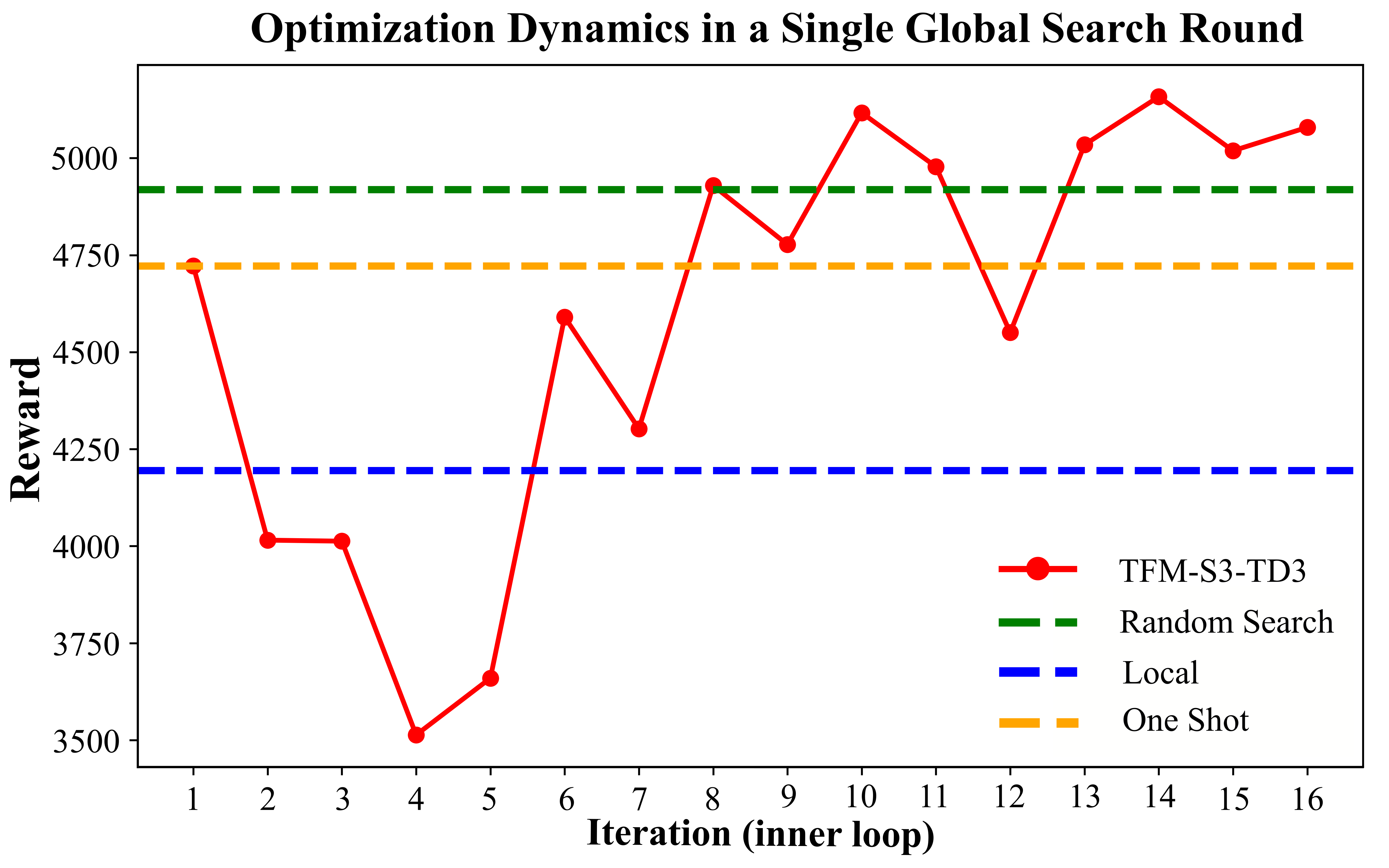}
\caption{
\textbf{Optimization dynamics within a global search round.}
The red curve shows the candidate return evaluated at each inner iteration (1–16).
The dashed green line denotes the best reward among the initial context set (16 sampled candidates) prior to search.
The dashed blue line indicates the policy reward obtained prior to the start of the global search.
The dashed yellow line represents the policy reward achieved after the first surrogate-guided iteration.
}
    \label{fig:single_round}
\end{figure}

To investigate whether \textbf{TFM-S3} consistently contributes to policy improvement throughout training, 
we visualize the improvement heatmap in Fig.~\ref{fig:heat_map}. 
Each row corresponds to a search round (grouped by training step), 
and each column represents one inner iteration. 
The color encodes the improvement value
\[
\Delta y = y_{\text{next}} - y_{0},
\]
where $y_{0}$ denotes the best reward among the initial context set (16 sampled candidates) before internal search, 
and $y_{\text{next}}$ is the reward obtained by the newly selected candidate at the current iteration. 
Positive values ($\Delta y > 0$) indicate that the candidate policy outperforms the round-specific baseline.

In the early rounds, the heatmap exhibits dense, highly saturated red regions corresponding to high $\Delta y$ values. Improvements are both frequent and large in magnitude, indicating that internal search is able to discover candidates that substantially outperform the baseline. This suggests that during early training, when the policy is still far from convergence, the search space contains many promising directions. The large-amplitude gains reflect strong exploration capability and significant improvement margins. In this phase, the global search does not merely provide incremental refinement; rather, it identifies substantially better candidates within each round.

In contrast, during later rounds, the red regions become noticeably lighter and less extensive. Typically, only two to three inner iterations per round produce positive improvements relative to the baseline. Improvements remain consistent, although of reduced magnitude. This pattern aligns with expected convergence dynamics: as the policy approaches a locally optimal region, the attainable improvement margin naturally shrinks. The reduced saturation therefore reflects the diminishing magnitude of the improvement rather than a loss of effectiveness. The internal search remains productive, but the gains are marginal.

Crucially, nearly every round across the entire training process contains at least one positive improvement relative to the baseline. Although the magnitude of gains decreases over time, the presence of improvement remains persistent. This demonstrates that internal search not only accelerates early learning but also provides beneficial refinement throughout training.

To better understand the optimization behavior within a single round, we visualize a representative search round in Fig.~\ref{fig:single_round}. Notably, improvements are not strictly monotonic. Certain iterations may produce temporary drops, reflecting exploration within the trust region.
However, the overall trend demonstrates clear upward refinement, and the best candidate found during the inner loop
substantially exceeds the pre-search policy reward
(blue dashed line). This illustrates that the surrogate-guided internal loop acts as a structured local optimizer inside the dynamically constructed subspace.

\subsection{Prediction Accuracy of the Tabular Foundation Model}
\label{sec:prediction_accuracy}

\begin{figure}[t]
    \centering
    \includegraphics[width=1.0\linewidth]{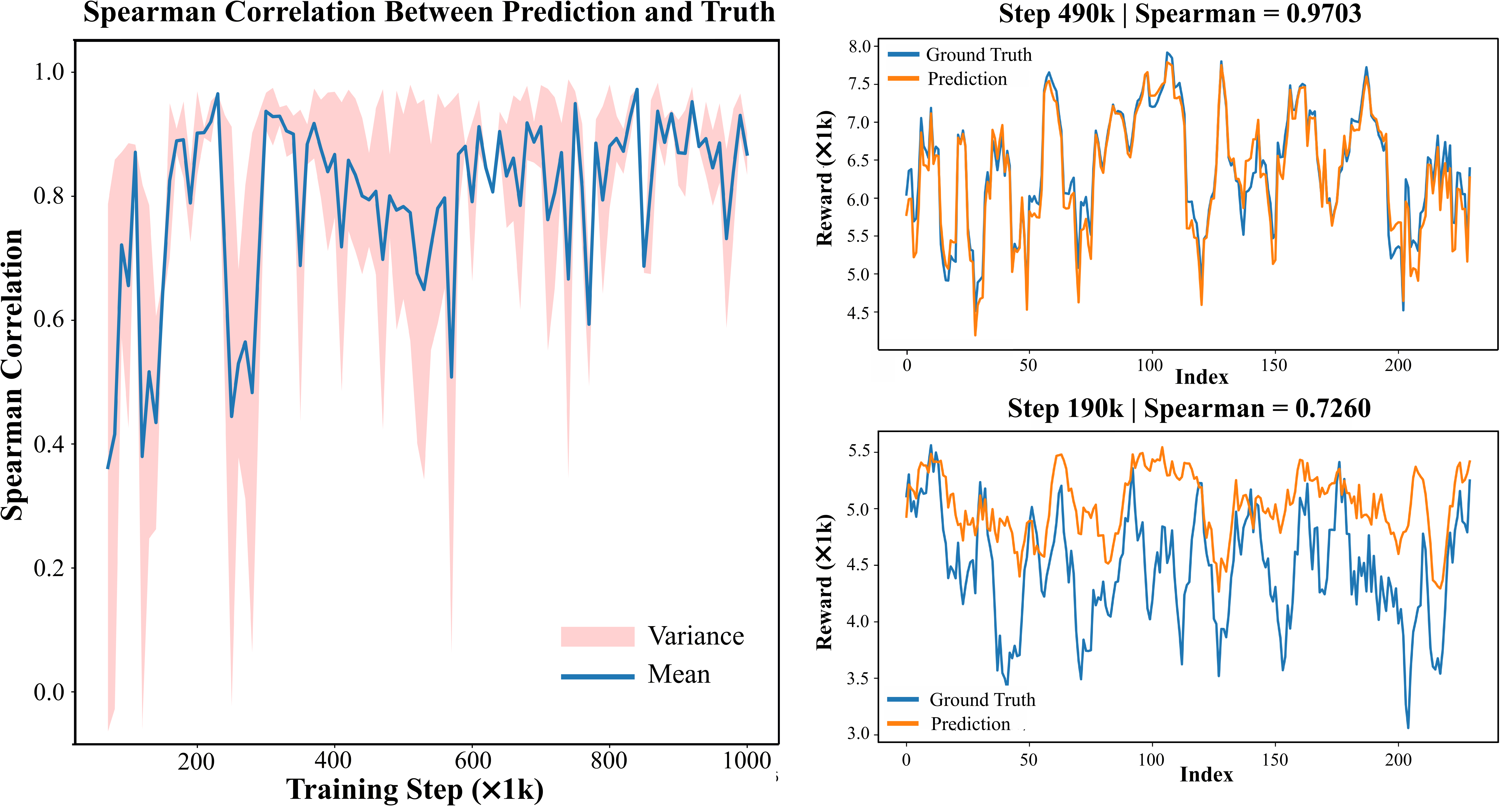}
\caption{
\textbf{Ranking consistency of the Tabular Foundation Model.}
\textbf{Left:} Spearman rank correlation between surrogate predictions and ground-truth returns across training steps (mean $\pm$ variance over 3 random seeds).
Spearman correlation measures agreement between predicted and true rankings.
After an initial unstable phase, the correlation increases and remains high for most of the training, indicating strong ranking consistency. This justifies the delayed activation of surrogate-guided subspace search. \\
\textbf{Right:} Representative comparisons at two stages of training.
At step 190k (bottom), agreement is moderate, with a Spearman correlation of 0.7260.
At step 490k (top), the predicted and ground-truth rankings align closely, with a Spearman correlation reaching 0.9703.
}
    \label{fig:pearson}
\end{figure}

\begin{table}[t]
\caption{\textbf{Quantitative evaluation of ranking consistency.}
We show the ground-truth percentile distribution of the surrogate-selected top-1 candidate ($N=256$).
}
\centering
\small
\begin{tabular}{c c c}
\toprule
\textbf{Percentile Range} & \textbf{TFM-S3 (\%)} & \textbf{Random (\%)} \\
\midrule
Top 0--20\%   & \textbf{63.5} & 20.0 \\
Top 20--40\%  & 28.2 & 20.0 \\
Top 40--100\% & 8.2  & 60.0 \\
\bottomrule
\end{tabular}
\label{top1_rank_distribution}
\end{table}

To assess the reliability of surrogate-guided screening, we analyze its behavior from two complementary perspectives:
(i) overall ranking consistency across the candidate pool, and 
(ii) decision quality of the surrogate-selected top-1 candidate.

\paragraph{Overall Ranking Consistency}
Our algorithm selects candidates according to their predicted ordering rather than their absolute predicted values. Consequently, ranking-based metrics are more informative than linear correlation measures.

We therefore adopt the Spearman rank correlation coefficient $\rho_s$ to quantify the monotonic agreement between surrogate predictions and ground-truth rollout returns. Formally,

\begin{equation}
\rho_s = \mathrm{corr}\big(r_x(i), r_y(i)\big),
\end{equation}

where $r_x(i)$ denotes the rank position of the predicted return $x_i$, and $r_y(i)$ denotes the rank position of the ground-truth return $y_i$. The pairs ${(x_i, y_i)}_{i=1}^{N}$ are computed within a candidate set of size $N=256$.

Figure~\ref{fig:pearson} (left) shows the Spearman correlation throughout training.
During early training, when the context set is limited and the policy landscape is highly non-stationary, ranking agreement is unstable.
As additional evaluations accumulate and the subspace geometry stabilizes, ranking consistency improves and remains high for the majority of training.

Figure~\ref{fig:pearson} (right) provides representative examples.
At step 190k, ranking alignment is moderate.
At step 490k, predicted and ground-truth rankings closely match, confirming reliable surrogate-based ordering of candidates.

\paragraph{Decision Quality of the Selected Candidate}
While Spearman measures global ranking agreement, our algorithm ultimately evaluates only the surrogate-based maximizer.
We therefore analyze the ground-truth percentile position of the predicted top-1 candidate.

Table~\ref{top1_rank_distribution} reports the distribution of the ground-truth percentile rank across all evaluation steps (over 3 random seeds).
Notably, 63.5\% of surrogate-selected candidates lie within the top 20\% of the true ranking, compared to only 20\% under random selection.
Moreover, over \textbf{91.8\%} (63.5 + 28.2) of selected candidates fall within the top 40\%, indicating that the surrogate consistently concentrates high-return policies near the top of the candidate pool.
Only 8.2\% of selected candidates lie outside the top 40\%, demonstrating that poor selections are relatively rare.

Taken together, these results show that the \ac{TFM} achieves strong overall ranking consistency and translates this consistency into high-quality decision-making.

This result highlights the potential of \ac{TFM} for robotics. Although the model observes only $32$ carefully selected samples within a subspace reduced from over $100\text{k}$ dimensions to $15$, its predictions remain sufficiently accurate to guide exploration toward high-reward regions.

\section{Conclusion}

We presented TFM-S3, a foundation-model–guided subspace-search framework for high-dimensional policy optimization in continuous control. By dynamically constructing a low-dimensional policy subspace from gradient statistics and performing surrogate-guided exploration within that subspace, our approach enables structured search beyond purely local gradient updates while avoiding exhaustive population-style rollout evaluation.

Across MuJoCo benchmarks, TFM-S3 consistently accelerates early-stage convergence and improves final performance under an identical rollout budget, while reducing variability across random seeds. These results indicate that foundation-model-based global search, when combined with dynamically evolving search geometry, provides an effective and interaction-efficient alternative to purely stochastic perturbation strategies.

The proposed framework serves as a general and modular enhancement to local, gradient-based policy optimization methods. 
In future work, we plan to integrate it with other actor–critic algorithms such as SAC and PPO, as well as with online optimal control approaches. 
Moreover, exploring alternative acquisition strategies and developing adaptive mechanisms for subspace dimension selection may further improve both efficiency and performance. 
More broadly, this work is evidence that TFMs can serve as expressive and adaptable data-driven priors in robotics, with a broad range of potential applications.


\begin{thebibliography}{10}
\providecommand{\url}[1]{#1}
\csname url@samestyle\endcsname
\providecommand{\newblock}{\relax}
\providecommand{\bibinfo}[2]{#2}
\providecommand{\BIBentrySTDinterwordspacing}{\spaceskip=0pt\relax}
\providecommand{\BIBentryALTinterwordstretchfactor}{4}
\providecommand{\BIBentryALTinterwordspacing}{\spaceskip=\fontdimen2\font plus
\BIBentryALTinterwordstretchfactor\fontdimen3\font minus \fontdimen4\font\relax}
\providecommand{\BIBforeignlanguage}[2]{{%
\expandafter\ifx\csname l@#1\endcsname\relax
\typeout{** WARNING: IEEEtran.bst: No hyphenation pattern has been}%
\typeout{** loaded for the language `#1'. Using the pattern for}%
\typeout{** the default language instead.}%
\else
\language=\csname l@#1\endcsname
\fi
#2}}
\providecommand{\BIBdecl}{\relax}
\BIBdecl

\bibitem{sutton1998reinforcement}
R.~S. Sutton, A.~G. Barto \emph{et~al.}, \emph{{Reinforcement learning: An introduction}}.\hskip 1em plus 0.5em minus 0.4em\relax MIT press Cambridge, 1998, vol.~1, no.~1.

\bibitem{schulman2017proximal}
J.~Schulman, F.~Wolski, P.~Dhariwal, A.~Radford, and O.~Klimov, ``{Proximal policy optimization algorithms},'' \emph{arXiv preprint arXiv:1707.06347}, 2017.

\bibitem{fujimoto2018addressing}
S.~Fujimoto, H.~Hoof, and D.~Meger, ``{Addressing function approximation error in actor-critic methods},'' in \emph{International Conference on Machine Learning}.\hskip 1em plus 0.5em minus 0.4em\relax PMLR, 2018, pp. 1587--1596.

\bibitem{duan2016benchmarking}
Y.~Duan, X.~Chen, R.~Houthooft, J.~Schulman, and P.~Abbeel, ``{Benchmarking deep reinforcement learning for continuous control},'' in \emph{International Conference on Machine Learning}.\hskip 1em plus 0.5em minus 0.4em\relax PMLR, 2016, pp. 1329--1338.

\bibitem{mnih2016asynchronous}
V.~Mnih, A.~P. Badia, M.~Mirza, A.~Graves, T.~Lillicrap, T.~Harley, D.~Silver, and K.~Kavukcuoglu, ``{Asynchronous methods for deep reinforcement learning},'' in \emph{International Conference on Machine Learning}.\hskip 1em plus 0.5em minus 0.4em\relax PMLR, 2016, pp. 1928--1937.

\bibitem{salimans2017evolution}
T.~Salimans, J.~Ho, X.~Chen, S.~Sidor, and I.~Sutskever, ``{Evolution strategies as a scalable alternative to reinforcement learning},'' \emph{arXiv preprint arXiv:1703.03864}, 2017.

\bibitem{snoek2012practical}
J.~Snoek, H.~Larochelle, and R.~P. Adams, ``{Practical bayesian optimization of machine learning algorithms},'' \emph{Advances in neural information processing systems}, vol.~25, 2012.

\bibitem{li2018measuring}
C.~Li, H.~Farkhoor, R.~Liu, and J.~Yosinski, ``{Measuring the Intrinsic Dimension of Objective Landscapes},'' in \emph{International Conference on Learning Representations}, 2018.

\bibitem{shahriari2015taking}
B.~Shahriari, K.~Swersky, Z.~Wang, R.~P. Adams, and N.~De~Freitas, ``{Taking the human out of the loop: A review of Bayesian optimization},'' \emph{Proceedings of the IEEE}, vol. 104, no.~1, pp. 148--175, 2015.

\bibitem{seeger2004gaussian}
M.~Seeger, ``{Gaussian processes for machine learning},'' \emph{International journal of neural systems}, vol.~14, no.~02, pp. 69--106, 2004.

\bibitem{hollmanntabpfn}
N.~Hollmann, S.~M{\"u}ller, K.~Eggensperger, and F.~Hutter, ``{TabPFN: A Transformer That Solves Small Tabular Classification Problems in a Second},'' in \emph{International Conference on Learning Representations}, 2023.

\bibitem{hussain2025foundation}
A.~Hussain, S.~Ali, U.~E. Farwa, M.~A.~I. Mozumder, and H.-C. Kim, ``{Foundation models: from current developments, challenges, and risks to future opportunities},'' in \emph{2025 27th International Conference on Advanced Communications Technology (ICACT)}.\hskip 1em plus 0.5em minus 0.4em\relax IEEE, 2025, pp. 51--58.

\bibitem{daniel2018tutorial}
J.~R. Daniel, V.~R. Benjamin, K.~Abbas, O.~Ian, and W.~Zheng, ``{A tutorial on thompson sampling},'' \emph{Foundations and Trends{\textregistered} in Machine Learning}, vol.~11, no.~1, pp. 1--99, 2018.

\bibitem{jaderberg2017population}
M.~Jaderberg, V.~Dalibard, S.~Osindero, W.~M. Czarnecki, J.~Donahue, A.~Razavi, O.~Vinyals, T.~Green, I.~Dunning, K.~Simonyan \emph{et~al.}, ``{Population based training of neural networks},'' \emph{arXiv preprint arXiv:1711.09846}, 2017.

\bibitem{conti2018improving}
E.~Conti, V.~Madhavan, F.~Petroski~Such, J.~Lehman, K.~Stanley, and J.~Clune, ``{Improving exploration in evolution strategies for deep reinforcement learning via a population of novelty-seeking agents},'' \emph{Advances in neural information processing systems}, vol.~31, 2018.

\bibitem{hansen2001completely}
N.~Hansen and A.~Ostermeier, ``{Completely derandomized self-adaptation in evolution strategies},'' \emph{IEEE Transactions on Evolutionary Computation}, vol.~9, no.~2, pp. 159--195, 2001.

\bibitem{omidvar2021review}
M.~N. Omidvar, X.~Li, and X.~Yao, ``{A review of population-based metaheuristics for large-scale black-box global optimization—Part I},'' \emph{IEEE Transactions on Evolutionary Computation}, vol.~26, no.~5, pp. 802--822, 2021.

\bibitem{eriksson2019scalable}
D.~Eriksson, M.~Pearce, J.~Gardner, R.~D. Turner, and M.~Poloczek, ``{Scalable global optimization via local Bayesian optimization},'' \emph{Advances in neural information processing systems}, vol.~32, 2019.

\bibitem{frazier2018bayesian}
P.~I. Frazier, ``{Bayesian optimization},'' in \emph{Recent advances in optimization and modeling of contemporary problems}.\hskip 1em plus 0.5em minus 0.4em\relax Informs, 2018, pp. 255--278.

\bibitem{wang2016bayesian}
Z.~Wang, F.~Hutter, M.~Zoghi, D.~Matheson, and N.~De~Feitas, ``{Bayesian optimization in a billion dimensions via random embeddings},'' \emph{Journal of Artificial Intelligence Research}, vol.~55, pp. 361--387, 2016.

\bibitem{auger2012tutorial}
A.~Auger and N.~Hansen, ``{Tutorial CMA-ES: evolution strategies and covariance matrix adaptation},'' in \emph{Proceedings of the 14th annual conference companion on Genetic and evolutionary computation}, 2012, pp. 827--848.

\bibitem{chengcontinuous}
Q.~Cheng, Y.~Wan, L.~Wu, C.~Hou, and L.~Zhang, ``{Continuous subspace optimization for continual learning},'' in \emph{Annual Conference on Neural Information Processing Systems}, 2025.

\bibitem{constantine2015active}
P.~G. Constantine, \emph{{Active subspaces: Emerging ideas for dimension reduction in parameter studies}}.\hskip 1em plus 0.5em minus 0.4em\relax SIAM, 2015.

\bibitem{hollmann2025accurate}
N.~Hollmann, S.~M{\"u}ller, L.~Purucker, A.~Krishnakumar, M.~K{\"o}rfer, S.~B. Hoo, R.~T. Schirrmeister, and F.~Hutter, ``{Accurate predictions on small data with a tabular foundation model},'' \emph{Nature}, vol. 637, no. 8045, pp. 319--326, 2025.

\bibitem{yugit}
R.~T.-Y. Yu, C.~Picard, and F.~Ahmed, ``{GIT-BO: High-Dimensional Bayesian Optimization with Tabular Foundation Models},'' in \emph{International Conference on Learning Representations}, 2026.

\end{thebibliography}
\end{document}